# Enabling Massage Actions: An Interactive Parallel Robot with Compliant Joints

Huixu Dong, Yue Feng, Chen Qiu, Ye Pan, Miaoying He, I-Ming Chen, Fellow, *IEEE*

*Abstract*—We propose a parallel massage robot with compliant joints based on the series elastic actuator (SEA), offering a unified force-position control approach. First, the kinematic and static force models are established for obtaining the corresponding control variables. Then, a novel force-position control strategy is proposed to separately control the force-position along the normal direction of the surface and another two-direction displacement, without the requirement of a robotic dynamics model. To evaluate its performance, we implement a series of robotic massage experiments. The results demonstrate that the proposed massage manipulator can successfully achieve desired forces and motion patterns of massage tasks, arriving at a high-score user experience.

## I. INTRODUCTION

Robotic manipulation technologies that are designed for domestic and workplace applications are engaging in physical activities [1-4], such as hands-on body massage [5]. Massage is in general a repetitive process, also a tedious, laborious, and time-consuming task. It is no surprise that an autonomous robot is an ideal worker that can replace a therapist to do the massage by initiating hands-on contact with humans.

Commercialized massage products, such as massage chairs and hammers, cannot usually reach users' satisfaction in their functional intelligence and prices [6, 7]. A two-end-effector massage device was developed to poke soft tissues [5], which cannot realize a compliant interaction. Jones *et al.* [8] presented a kneading robot for conducting the ring and linear massaging actions through an industrial robot, and Huang *et al.* developed a robot equipped with four elastic joints [9]; however, these robotic systems are expensive. An oral rehabilitation robot was applied to perform the rotational movement to rub the maxillofacial tissues of the face [10]. This device just has a single massage function. Kang *et al.* proposed a conceptual mechanism design for the robot implementing a tapping massage task, without the further implementation of a control strategy [11]. Some compliant mechanisms in interactive robots regulate the contact forces and conduct adaptable behaviors with minimal control effort [12-14]. Compliant control has been thoroughly reviewed concerning the main two basic control such as hybrid force-position control and impedance control [15]. In addition, the emerging techniques promote the functional intelligence of massage robots. The point cloud was used for generating body trajectory for robot massage applications [16]. Si *et al.* allowed a massage robot to be more intelligent via the green Internet of Things [17]. For existing techniques, the high complexity, considerable cost and heavy structure of the mechatronics system result in the limitation of widespread applications [18].

To tackle the above problems, we develop a robotic manipulator which aims to assist physicians and conduct soft-tissue massage treatments for patients by the control of compliant movement to remain in contact with the interactive force (see Fig.1). The proposed massage robot can provide as follows: 1) it is convenient for this massage robot to be deployed at home or small clinics due to its compact and space-saving structure; 2) stable versatile massage treatments include pressing, tapping, rolling and pushing. The proposed parallel manipulator is made up of three symmetric fingers with proximal links, parallelogram links, and distal links connecting from the base to the end effector in parallel, which allows three translational motions and generally no rotational movement of the end effector [19]. Moreover, to guarantee safety, a rotary compliant joint with a series elastic actuator is designed and then, integrated into the manipulator; 3) safe and efficient massage treatment is implemented by a compliant control strategy. After constructing kinematic and static force models of the proposed message robot, we design a control method that tracks force and position of a manipulator end-effector based on electronic encoders, which does not require the dynamics model and force sensor.

## II. METHODOLOGY

### A. Robot Kinematics Modelling

*1) The geometry of the proposed massage robot*

The kinematic chain of the proposed parallel manipulator is illustrated in Fig. 2. The fixed base and the end-effector Cartesian reference frame are denoted as $\{B\}$ and $\{P\}$, whose origins locate in their centers, respectively. In addition, the orientations of $\{B\}$ and $\{P\}$ are the same and thus, the rotation matrix $^B_P R = I_3$ is constant. We have the active joint variables $\boldsymbol{\theta} = \{\theta_1, \theta_2, \theta_3\}^T$ and the position variables

Huixu Dong is with Robot Perception and Grasp Lab, Zhejiang University, 310027 China; Yue Feng, I-Ming Chen are with Robotics Research Center, Nanyang Technological University, 639798 Singapore; Chen Qiu is with Maider Medical Industry Equipment Co., ltd, 317607 China; Ye Pan is with Shanghai Jiaotong University, 200240 China; Miaoying He is with Chongqing University of Technology, 400054, China.

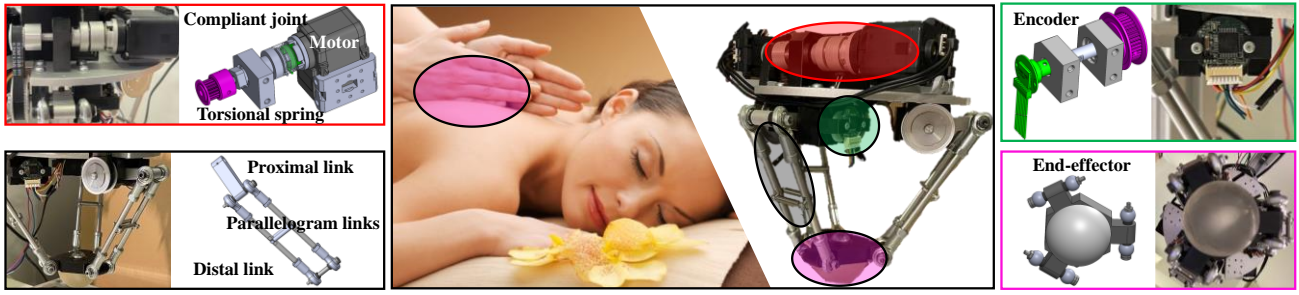

Figure. 1. The design of the proposed parallel massage manipulator consisting of compliant joints, encoders, links and end-effector. A potential application scenario where the proposed massage robot can perform a tapping massage task (Middle figure).

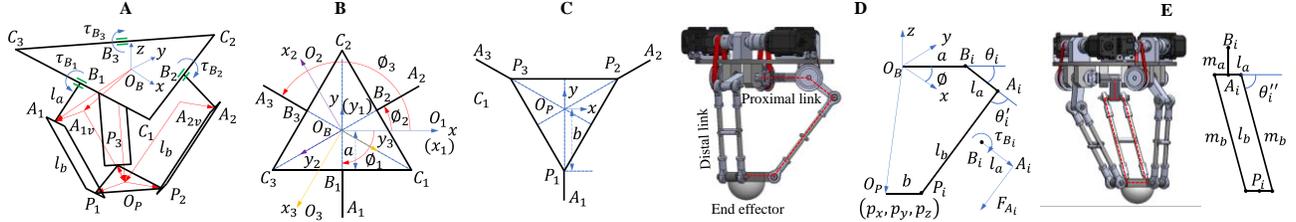

Figure. 2. Geometry of the proposed parallel manipulator. The whole sketch(A); the fix-based platform with the coordinate reference frame $O_B$(B); the end-effector moving platform with the coordinate reference frame $O_P$ (C); the two views of the manipulator (D and E). The green lines represent the actively rotational joints(A). $x_i, y_i, z_i$ denote the coordinate reference frame and the corresponding axis $x_i$ is parallel to the direction of the rational joints (the green lines in A) at $B_i$ ($i = 1, 2, 3$)(B). $m_a$ and $m_b$ represent the masses of the proximal rod (link) and distal rod (two rod consisting of link), respectively. $l_a$ and $l_b$ denote the proximal and distal link lengths, respectively.

$\{p_x, p_y, p_z\}^T$ of the end effector. To achieve the relationship between active joint and position variables, we build the following equation according to the vector-loop diagram with the kinematic parameters,

$$\overrightarrow{O_B B_i} + \overrightarrow{B_i A_i} + \overrightarrow{A_i P_i} = \overrightarrow{O_B O_P} + \overrightarrow{O_P P_i} \quad (1)$$

for $i = 1, 2, 3$, and thus, the geometrical constraints based on the distal link length can be described by

$$l_b = \|\overrightarrow{A_i P_i}\| = \|\overrightarrow{O_B O_P} + \overrightarrow{O_P P_i} - \overrightarrow{O_B B_i} - \overrightarrow{B_i A_i}\| \quad (2)$$

As shown in Fig. 2, $B_i$ of the fixed-base revolute joint point in {B} and $P_i$ of the end-effector U-joint connection point in {P} can be given as

$$^B B_1 = \{0 \ -a \ 0\}^T,$$
$$^B B_2 = \left\{\tfrac{\sqrt{3}}{2}a \ \tfrac{1}{2}a \ 0\right\}^T,$$
$$^B B_3 = \left\{-\tfrac{\sqrt{3}}{2}a \ \tfrac{1}{2}a \ 0\right\}^T, \quad (3)$$

$$^P P_1 = \{0 \ -b \ 0\}^T,$$
$$^P P_2 = \left\{\tfrac{\sqrt{3}}{2}b \ \tfrac{1}{2}b \ 0\right\}^T,$$
$$^P P_3 = \left\{-\tfrac{\sqrt{3}}{2}b \ \tfrac{1}{2}b \ 0\right\}^T, \quad (4)$$

According to the active joint variables $\{\theta_1, \theta_2, \theta_3\}^T$, we can obtain the vector $\overrightarrow{B_i A_i}$ of the fixed-base revolute joint point in {B}

$$^B \overrightarrow{B_1 A_1} = \{0 \ -l_a \cos \theta_1 \ -l_a \sin \theta_1\}^T,$$
$$^B \overrightarrow{B_2 A_2} = \{l_a \cos \theta_2 \cos \emptyset_2 \ l_a \cos \theta_2 \sin \emptyset_2 \ -l_a \sin \theta_2\}^T,$$
$$^B \overrightarrow{B_3 A_3} = \{-l_a \cos \theta_3 \cos \emptyset_3 \ l_a \cos \theta_3 \sin \emptyset_3 \ -l_a \sin \theta_3\}^T, \quad (5)$$

where $\emptyset_i = -\tfrac{\pi}{2} + \tfrac{2\pi}{3}(i-1)$, $i = 1, 2, 3$; $a$ and $b$ represent the radii of the fixed and end-effector platform, respectively; $l_a$ and $l_b$ are the proximal/upper and distal/lower link lengths, respectively; $\emptyset_i$ denotes the rotation angle concerning the $z$ axis for the point $B_i$ in {B}; the actuated joint angle is represented by $\theta_i$ and passive joint angles are indicated by $\theta_i'$ and $\theta_i''$ (see Fig. 2). Thus, we have $\overrightarrow{A_i P_i}$ as follows

$$^B \overrightarrow{A_1 P_1} = \{p_x \ p_y + l_a \cos \theta_1 + k \ p_z + l_a \sin \theta_1\}^T,$$
$$^B \overrightarrow{A_2 P_2} = \begin{Bmatrix} p_x - l_a \cos \theta_2 \cos \emptyset_2 + m \\ p_y - l_a \cos \theta_2 \sin \emptyset_2 + n \\ p_z + l_a \sin \theta_2 \end{Bmatrix}$$
$$^B \overrightarrow{A_3 P_3} = \begin{Bmatrix} p_x + l_a \cos \theta_3 \cos \emptyset_3 - m \\ p_y - l_a \cos \theta_3 \sin \emptyset_3 + n \\ p_z + l_a \sin \theta_3 \end{Bmatrix} \quad (6)$$

in which $k = a - b$, $m = \tfrac{\sqrt{3}}{2}b - \tfrac{\sqrt{3}}{2}a$, $n = \tfrac{1}{2}b - \tfrac{1}{2}a$.

The following kinematics equations for the proposed manipulator are obtained as

$$\|\overrightarrow{A_i P_i}\|^2 - l_b^2 = 0 \quad (7)$$

which can be further utilized to conduct forward and inverse position kinematics analysis.

*2) Inverse position kinematics(IPK)*

In terms of inverse position kinematics, the Cartesian position $\{p_x, p_y, p_z\}^T$ of the end-effector is provided to calculate the three actuated revolute joint angles $\{\theta_1, \theta_2, \theta_3\}^T$. Referring to Eq.(7), it can be further described as

$$E_i \cos \theta_i + F_i \sin \theta_i + G_i = 0 \quad (8)$$

which can be addressed through the tangent half-angle substitution. If we define

$$t_i = \tan \tfrac{\theta_i}{2}, \ \cos \theta_i = \tfrac{1-t_i^2}{1+t_i^2}, \ \cos \theta_i = \tfrac{2 t_i}{1+t_i^2}, \quad (9)$$

then Eq.(8) can be rearranged as

$$(G_i - E_i) t_i^2 + (2 F_i) t_i + (G_i + E_i) = 0 \quad (10)$$

If $E_i^2 + F_i^2 > G_i^2$, we can obtain

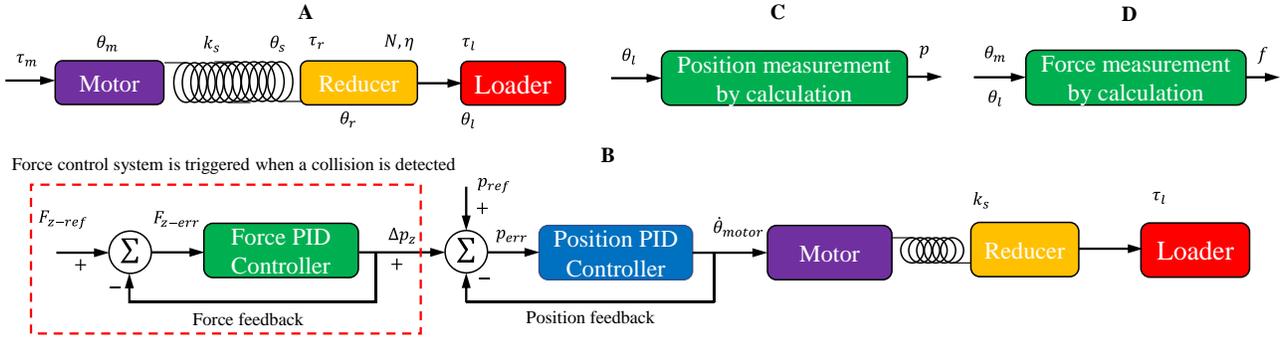

Figure. 3. The diagram of the transmission model of the compliant joint(A), the overview of the compliant actuator system(B), the force(C) and position(D) measurements by calculations. $F_{z-ref}$ and $p_{ref}$ denote the z-axis pre-defined force and the pre-defined position $(p_x, p_y, p_z)$, respectively; $\Delta p_z$ represents the z-axis changed position determined by the force control system; $F_{z-err}$ and $p_{err}$ indicate the z-axis force error and the position error, respectively; $\dot{\theta}_{motor}$ is the velocity of the motor.

$$t_{i_{1,2}} = \frac{-F_i \pm \sqrt{E_i^2 + F_i^2 - G_i^2}}{G_i - E_i} \quad (11)$$

and

$$\theta_i = 2\tan^{-1}(t_i) \quad (12)$$

Note that two $\theta_i$ are both valid solutions.

*3) Forward position kinematics(FPK)*

Given the three actuated joint angles $\{\theta_1, \theta_2, \theta_3\}^T$, the resulting Cartesian position of the end-effector moving platform is obtained. According to the vector loop, we have

$$\overrightarrow{O_B B_i} + \overrightarrow{B_i A_i} = \overrightarrow{O_B A_i} \quad (13)$$

which are

$${}^B\overrightarrow{O_B A_1} = \{0 \quad -a - l_a\cos\theta_1 \quad -l_a\sin\theta_1\}^T$$

$${}^B\overrightarrow{O_B A_2} = \begin{Bmatrix} \frac{\sqrt{3}}{2}(a + l_a\cos\theta_2) \\ \frac{1}{2}(a + l_a\cos\theta_2) \\ -l_a\sin\theta_2 \end{Bmatrix}$$

$${}^B\overrightarrow{O_B A_3} = \begin{Bmatrix} -\frac{\sqrt{3}}{2}(a + l_a\cos\theta_3) \\ \frac{1}{2}(a + l_a\cos\theta_3) \\ -l_a\sin\theta_3 \end{Bmatrix} \quad (14)$$

If we define three virtual sphere centers based on ${}^B_P R = I_3$ through

$$\overrightarrow{O_B A_i} - \overrightarrow{O_P P_i} = \overrightarrow{O_B A_{iv}} \quad (15)$$

$${}^B\overrightarrow{O_B A_{1v}} = \{\overline{OA}_{1x} \quad \overline{OA}_{1y} \quad \overline{OA}_{1z}\}^T$$
$${}^B\overrightarrow{O_B A_{2v}} = \{\overline{OA}_{2x} \quad \overline{OA}_{2y} \quad \overline{OA}_{2z}\}^T$$
$${}^B\overrightarrow{O_B A_{3v}} = \{\overline{OA}_{3x} \quad \overline{OA}_{3y} \quad \overline{OA}_{3z}\}^T$$

$\overline{OA}_{1x} = 0,$
$\overline{OA}_{1y} = -a - l_a\cos\theta_1 + b,$
$\overline{OA}_{1z} = -l_a\sin\theta_1,$
$\overline{OA}_{2x} = \frac{\sqrt{3}}{2}(a + l_a\cos\theta_2) - \frac{\sqrt{3}}{2}b,$
$\overline{OA}_{2y} = \frac{1}{2}(a + l_a\cos\theta_2) - \frac{1}{2}b,$
$\overline{OA}_{2z} = -l_a\sin\theta_2,$
$\overline{OA}_{3x} = -\frac{\sqrt{3}}{2}(a + l_a\cos\theta_3) + \frac{\sqrt{3}}{2}b,$
$\overline{OA}_{3y} = \frac{1}{2}(a + l_a\cos\theta_3) - \frac{1}{2}b,$
$\overline{OA}_{3z} = -l_a\sin\theta_3,$

then the FPK solution is the intersection point of three known spheres. Let a sphere be referred as a vector $(\{c\}, r)$ with the center $\{c\}$ and scalar radius $r$. Thus, the FPK unknown point is the intersection of the three known spheres $(\{{}^B A_{iv}\}, l_b)$. Assume the three given spheres are $(c_i, l_b)$ where $c_i = \{x_i, y_i, z_i\}^T$ and radii $l_b$ are known. The equations of the three spheres are provided as

$$(x - x_i)^2 + (y - y_i)^2 + (z - z_i)^2 = l_b^2 \quad (16)$$

Then, we expand the equations above and subtract the third from the first and the third from the second, yielding

$$\begin{aligned} x &= f(y) \\ z &= f(y) \end{aligned} \quad (17)$$

These two functions are substituted into the sphere equation for obtaining the position of the moving end-effector.

### B. Static Force Analysis

*1) The transmission model of compliant joint*

Figure 3(A) illustrates the transmission of the compliant joint based on the series elastic actuator (SEA). This system consists of a motor, a synchronous belt that amplifies the output force (torque) of the actuator, a spring placed between the motor and a synchronous belt, and the loader (parallel manipulator). To simplify the dynamic model, we consider the compliant joint based on SEA as an undamped linear torsion spring. As shown in Fig. 3(A), through the spring, the torques can be formulated as

$$\tau_m = \tau_r = k_s\theta_s = \frac{\tau_l}{N\eta} \quad (18)$$

where $\tau_m$ is the output torque of the motor(the input torque of the compliant element-torsion spring); $\tau_r$ denotes the output torque of the torsion spring; $k_s$ is the sprint stiffness; $N = 2$ is the reduction ratio of the synchronous belt and $\eta$ is the transmission efficiency, respectively; $\theta_s$ represents the changed angle from the input angle $\theta_m$ to the output angle $\theta_r$ of the compliant element, which is given as

$$\theta_s = \theta_m - \theta_r = \theta_m - N\theta_l \quad (19)$$

in which $\theta_l$ represents the angle of the load. Combining Eq.(18) and Eq.(19), we have

$$\tau_l = k_s N\eta(\theta_m - N\theta_l) \quad (20)$$

*2) The resultant force of the end-effector*

After achieving the end-effector position $\{p_x, p_y, p_z\}^T$ by the above algorithm, we do the static analysis to calculate the resultant force of the end-effector.

To obtain the unit vector of $\overrightarrow{A_i P_i}(i = 1,2,3)$ in Eq.(6), we calculate the magnitude $\|\overrightarrow{A_i P_i}\|$ of $\overrightarrow{A_i P_i}$ with the coordinate reference frame $O_B$ as follows,

$$\|{}^B\overrightarrow{A_i P_i}\| = \sqrt{X^2 + Y^2 + Z^2} \quad (21)$$

where $X, Y, Z$ represent the three components of a vector. Thus, the unit vector $\|{}^B_e\overrightarrow{A_i P_i}\|$ is given as

$$ {}^B_e\overrightarrow{A_i P_i} = \frac{{}^B\overrightarrow{A_i P_i}}{\|{}^B\overrightarrow{A_i P_i}\|} \quad (22)$$

To allow the coordinate of the unit vector of $\overrightarrow{A_i P_i}$ to be consistent with that of the loading torque at each proximal link, we need to respectively convert three unit vectors $\{{}^B_e\overrightarrow{A_i P_i}, {}^B_e\overrightarrow{A_i P_i}, {}^B_e\overrightarrow{A_i P_i}\}$ in the coordinate reference frame $O_B$ to the coordinate reference frames $\{O_1 - (x_1, y_1, z_1), O_2 - (x_2, y_2, z_2), O_3 - (x_3, y_3, z_3)\}$, as shown in Fig. 2(B). Note that the x-axes of these three coordinate frames are parallel to the rational axes of three proximal likes on the base. In particular, the coordinate reference frame $O_1$ coincides with the base coordinate reference frame $O_B$; $O_2$ is the frame that $O_B$ is counterclockwise rotated $\frac{2}{3}\pi$ to form; $O_B$ is clockwise rotated $\frac{2}{3}\pi$ to obtain $O_3$. Further, we use the two rotation matrices $R^z_{B-2}$ and $R^z_{B-3}$ to convert the unit vectors ${}^B_e\overrightarrow{A_2 P_2}$ and ${}^B_e\overrightarrow{A_3 P_3}$ to the coordinate reference frames $O_2$ and $O_3$, respectively. The rotation matrices $R^z_{B-2}$ and $R^z_{B-3}$ are provided as

$$R^z_{B-2} = \begin{bmatrix} \cos\left(\frac{2}{3}\pi\right) & -\sin\left(\frac{2}{3}\pi\right) & 0 \\ \sin\left(\frac{2}{3}\pi\right) & \cos\left(\frac{2}{3}\pi\right) & 0 \\ 0 & 0 & 1 \end{bmatrix}$$

$$R^z_{B-3} = \begin{bmatrix} \cos\left(-\frac{2}{3}\pi\right) & -\sin\left(-\frac{2}{3}\pi\right) & 0 \\ \sin\left(-\frac{2}{3}\pi\right) & \cos\left(-\frac{2}{3}\pi\right) & 0 \\ 0 & 0 & 1 \end{bmatrix} \quad (23)$$

Therefore, the unit vectors ${}^1_e\overrightarrow{A_1 P_1}, {}^2_e\overrightarrow{A_2 P_2}$ and ${}^3_e\overrightarrow{A_3 P_3}$ can be calculated by

$$\begin{aligned} {}^1_e\overrightarrow{A_1 P_1} &= {}^B_e\overrightarrow{A_1 P_1} \\ {}^2_e\overrightarrow{A_2 P_2} &= R^z_{B-2} \cdot {}^B_e\overrightarrow{A_2 P_2} \\ {}^3_e\overrightarrow{A_3 P_3} &= R^z_{B-3} \cdot {}^B_e\overrightarrow{A_3 P_3} \end{aligned} \quad (24)$$

In the 3D space, the torque is given by the cross product of the position vector(distance vector) and the force vector, such as $\tau = \vec{r} \times \vec{F}$ ($\tau$ representing the torque, $\vec{r}$ indicating the distance vector as well as $\vec{F}$ denoting the force vector. Employing this method, we can calculate the force magnitudes $F_{A_1}, F_{A_2}, F_{A_3}$ acting on points $A_1, A_2, A_3$ in the coordinate reference frames $\{O_1, O_2, O_3\}$, respectively. In particular, we have the following equations as

$$\begin{aligned} l_a \times F_{A_1} \cdot {}^1_e\overrightarrow{A_1 P_1} &= \tau_{B_1} \\ l_a \times F_{A_2} \cdot {}^2_e\overrightarrow{A_2 P_2} &= \tau_{B_2} \\ l_a \times F_{A_3} \cdot {}^3_e\overrightarrow{A_3 P_3} &= \tau_{B_3} \end{aligned} \quad (25)$$

where $\tau_{B_i}(i = 1,2,3)$ represents the loading torque at the rotation joint $B_i$. The value of $\tau_{B_i}$ along the x-axis of the coordinate reference frame $O_i$ is known, which is equal to $\tau_l$ for each motor. There is just one unknown force magnitude for one item of Eq.(22) and thus, we can obtain these force magnitudes. After that, $F_{A_2}$ in $O_2$ and $F_{A_3}$ in $O_3$ need to be converted to the coordinate reference frame $O_B(O_1)$ by the matrices $R^z_{B-3}$ and $R^z_{B-2}$ as follows,

$$\begin{aligned} {}^B F_{A_1} &= F_{A_1} \\ {}^B F_{A_2} &= R^z_{B-3} F_{A_2} \\ {}^B F_{A_3} &= R^z_{B-2} F_{A_3} \end{aligned} \quad (26)$$

in which ${}^B F_{A_i}(i = 1,2,3)$ indicates the force vector along $A_i P_i$ in the coordinate reference frame $O_B$. Finally, the resultant force ${}^B F_{ee}$ of the end-effector can be provided as

$${}^B F_{ee} = {}^B F_{A_1} + {}^B F_{A_2} + {}^B F_{A_3} \quad (27)$$

## C. Control System Design of the Robotic Manipulator

In practical applications, the massage manipulator has a relatively small workspace and small motion speed. Moreover, the proximal and distal links are relatively light. Thus, we implement the control strategy of the parallel massage manipulator without a dynamic model of the robot.

To achieve such a target of the manipulator performing massage tasks, we design a control system including the rear-loop position and the front-loop force, as shown in Fig. 3(B). For each loop, we adopt the PID feedback control strategy. As illustrated in Fig. 3(C, D), the built-in encoder of a motor and the encoder mounted on the rotation shaft of the proximal link can provide $\theta_m$ and $\theta_l$, respectively. To perform the path planning of the end-effector, its position can be obtained by the forward kinematics (see Fig. 3-C) at the corresponding time sample when the angle $\theta_l$ of the load is known and then, the static force analysis can be applied to determine the force of the end-effector. Further, we put $\theta_m$ and $\theta_l$ into Eq.(20) to obtain $\tau_l$ for calculating the end-effector's force. In terms of massage tasks, the robotic manipulator generally needs to guarantee that the force exerted on the skin tissue of the human body is constant along the z-axis direction.

## III. EXPERIMENTS

### A) Evaluations of the Massage System

In Eq.(20), $k_s$ and $\eta$ need to be confirmed for calculating the force of the end-effector. An electronic scale with a resolution of 0.005kg below the massage robot is considered as a force sensor to collect the contacting forces from the manipulator's end-effector. We allow the massage manipulator to touch the surface of the electronic scale, gradually increasing the z-axis position. Comparing the force calculated via the proposed static force model with the force measured by the electronic scale, we can obtain the errors of the calculated force in practical applications for adjusting the spring stiffness and the transmission efficiency, thus determining $k_s = 1, \eta = 1$.

*1) Evaluation of hybrid force & position control*

For evaluating the performance of the proposed massage manipulator on the proposed force-position control, we conducted two groups of comparison experiments on the same experimental platform, including the pure position control and hybrid force&position control (force-position control). For the setup of position control, the manipulator was allowed to move along the y-axis from $p_y = 20$ to $p_y = -20mm$ with $p_x = 0$ and $p_z = -105mm$ at a constant speed. Since the surface of the electronic scale is at $p_z = -101mm$, the manipulator generates a pressing force $F_z$ in the z-axis direction and frictional forces in the x-axis and y-axis directions. The end-effector has surface contact with the electronic scale, which results in the x-axis and y-axis frictions. In terms of force-position control, based on the above setup, we enable the force control loop to be triggered when $F_z$ is bigger than $0.6\ N$ and set a reference force $F_{z-ref} = 5N$ in the force control loop (see Fig. 4).

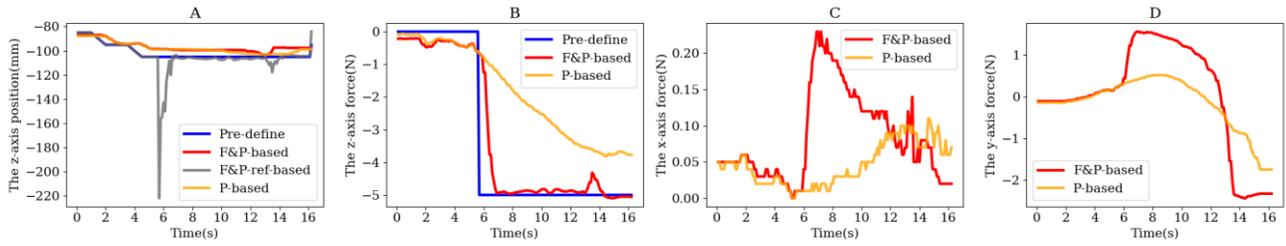

Figure. 4. The z-axis position (A), the z-axis forces (B), the x-axis forces(C) and the y-axis forces(D) at the time samples. The blue, yellow, orange curves represent the pre-defined trajectory or force as a reference value (Pre-define), the trajectory or force based on the pure position control(P-based), the trajectory or force based on hybrid force&position control(F&P-based), respectively. The grey curve indicates the z-axis pre-defined reference trajectory for hybrid force&position control(F&P-based-ref)(A). For the grey curve, initially, its z-axis trajectory overlaps the pre-define trajectory represented by the blue curve. During this period, the force control is not triggered since the pressing force is smaller than the set triggered threshold of 0.6N. Next, with the z-axis force increasing, the force control is triggered at around 5.6s. The z-axis position generates a sudden change since the z-axis changing position $\Delta p_z$ is involved in this pre-defined trajectory(see Fig.4-B). Further, this pre-defined trajectory for force-position control still goes back and almost overlaps the pre-defined trajectory.

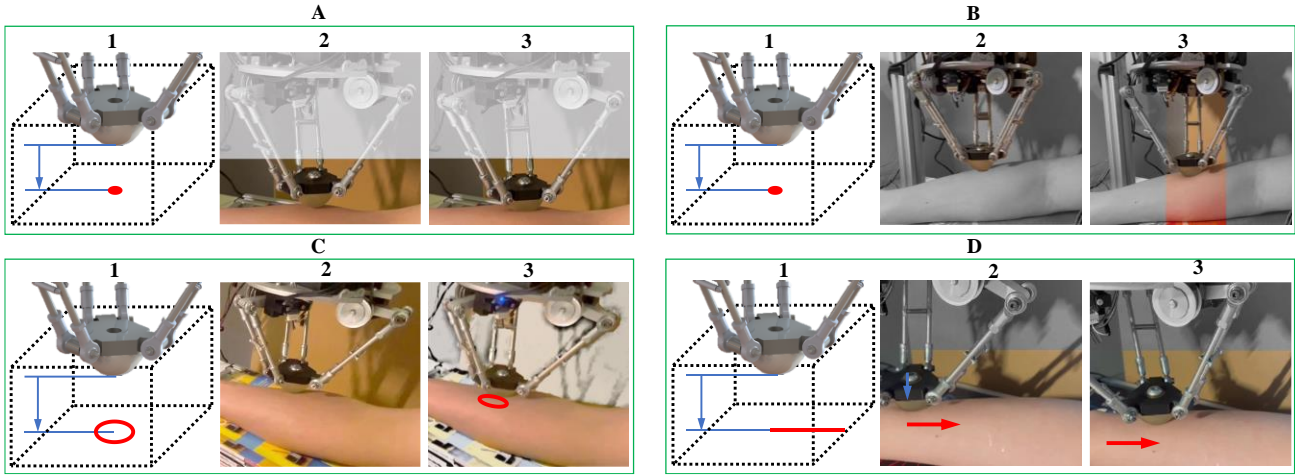

Figure. 5. The pressing(A), tapping(B), rolling(C) and pushing(D) massage tasks. The positions of the pressing and tapping massage tasks(A-1, B-1); the trajectories of the rolling and pushing massage tasks(C-1, D-1); the snapshots of the massage tasks(A-1-2, B-1-2, C-1-2, D-1-2). The black cubic denotes the workspace and the blue arrow indicates the moving direction of the end-effector. The red tiny circle and the red line represent the trajectories of the rolling and pushing massage tasks, respectively(C, D).

Figure 4(A) illustrates two scenarios where the end-effector follows the pre-defined trajectory and force under the pure position control and force-position control strategies. First, in terms of position control, the pre-defined (blue curve) and position-control (yellow curve) trajectories have similar changing trends and the average tracking difference of this trajectory keeps small values. This indicates that the used position control can enable the end-effector to accurately follow the pre-defined trajectory. The position tracking difference is generated owing to the mechanical backlash. Second, we evaluate the proposed force-position strategy. For the electronic scale, the measured force is in proportion to the z-axis displacement. That is, if the z-axis position is a constant, the applied force on the surface of the electronic scale remains unchanged. It is found that the z-axis trajectory based on the force-position control strategy is almost the same as the z-axis control-based trajectory. Thus, the proposed force-position control strategy can enable the end-effector to track the pre-defined trajectory and force.

The z-axis position control-based force $F_{z-p}$ and the force-position control-based $F_{z-fp}$ are calculated by the proposed method of the static force analysis, as illustrated in Fig. 4(B). The pre-defined force is considered as a reference force in the force control loop. While tracking, $F_{z-p}$ also increases as the end-effector needs to arrive at the pre-defined z-axis position; however, the force tracking based on the position control has unsatisfied performance such as a severe delay and big force error. Contrastively, the tracking performance of $F_{z-fp}$ is superior to that of $F_{z-p}$ owing to the involved force control. Although a small force fluctuation occurs at 13.7s, $F_{z-fp}$ can still successfully continue tracking the reference force.

The results of forces in the x-axis and y-axis directions are obtained by the proposed static force model and encoders, as illustrated in Fig. 4(C, D). The forces in the x-axis direction have small fluctuations compared to those in the y-axis direction. For the x-axis forces, this is somewhat expected since there are just frictional forces, without the x-axis force control(see Fig. 4-C). Moreover, since $F_{z-p}$ is smaller than $F_{z-fp}$, the frictional force based on the position control is less than one based on the force-position control (see Fig. 4-C, D).

*B) Robot Massage Tasks*

A series of experiments were carried out for investigating the massage effectiveness of the proposed manipulator (see Fig. 5), which is provided as follows:

· **Pressing**: the end-effector moves to a destination which is generally an acupuncture point, goes down gently, gradually increases the force until the pre-defined force is reached, and remains stationary for a while with this force.

· **Taping**: the robotic end-effector moves to a pre-taping position above the destination, which is generally an acupuncture point, goes down rapidly with an adjustable force until the pre-defined force is reached.

· **Rolling**: the robot end-effector moves to a destination which is generally an acupuncture point, goes down gently, gradually increases the force until the pre-defined force arrives, and then, moves periodically in the pattern of the tiny circle on the point with a fixed downward force.

· **Pushing**: the robotic end-effector moves to a destination which is generally an acupuncture point, goes down slowly with an adjustable force, then it pushes straight along the pre-defined trajectory with a fixed downward force.

We gathered feedback from their experiences. Overall, users' experience with the proposed robot surpassed their expectations, as reflected in their comments: "I got very comfortable with it". The proposed massage manipulator can successfully implement multiple massage tasks on humans, which is potentially employed in small clinics and household scenarios.

## IV. CONCLUSION

In this work, we established a novel parallel massage manipulator that can implement multiple-functional massage tasks and be potentially deployed in small clinics and households due to the suitable structure and low cost. Various experiments were performed to demonstrate that the proposed massage robot could successfully provide treatment aids for physicians.

ACKNOWLEDGEMENT

This work is partially supported by Science Fund for Creative Research Groups of National Natural Science Foundation of China (No.51821093).